# Face Expression Recognition and Analysis: The State of the Art

Vinay Bettadapura

College of Computing, Georgia Institute of Technology

vinay@gatech.edu

**Abstract** — The automatic recognition of facial expressions has been an active research topic since the early nineties. There have been several advances in the past few years in terms of face detection and tracking, feature extraction mechanisms and the techniques used for expression classification. This paper surveys some of the published work since 2001 till date. The paper presents a time-line view of the advances made in this field, the applications of automatic face expression recognizers, the characteristics of an ideal system, the databases that have been used and the advances made in terms of their standardization and a detailed summary of the state of the art. The paper also discusses facial parameterization using FACS Action Units (AUs) and MPEG-4 Facial Animation Parameters (FAPs) and the recent advances in face detection, tracking and feature extraction methods. Notes have also been presented on emotions, expressions and facial features, discussion on the six prototypic expressions and the recent studies on expression classifiers. The paper ends with a note on the challenges and the future work. This paper has been written in a tutorial style with the intention of helping students and researchers who are new to this field.

**Index Terms** — Expression recognition, emotion classification, face detection, face tracking, facial action encoding, survey, tutorial, human-centered computing.

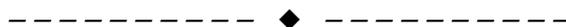

## 1. Introduction

Studies on Facial Expressions and Physiognomy date back to the early Aristotelian era (4th century BC). Physiognomy is the assessment of a person's character or personality from their outer appearance, especially the face [1]. But over the years, while the interest in Physiognomy has been waxing and waning[1] [1], the study of facial expressions has consistently been an active topic. The foundational studies on facial expressions that have formed the basis of today's research can be traced back to the 17th century. A detailed note on the various expressions and movement of head muscles was given in 1649 by John Bulwer in his book *"Pathomyotomia"*. Another interesting work on facial expressions (and Physiognomy) was by Le Brun, the French academician and painter. In 1667, Le Brun gave a lecture at the Royal Academy of Painting which was later reproduced as a book in 1734 [2]. It is interesting to know that the 18th century actors and artists referred to his book in order to achieve *"the perfect imitation of 'genuine' facial expressions"* [2][2]. The interested reader can refer to a recent work by J. Montagu on the origin and influence of Le Brun's lectures [3].

Moving on to the 19th century, one of the important works on facial expression analysis that has a direct relationship to the modern day science of automatic facial expression recognition was the work done by Charles Darwin. In 1872, Darwin wrote a treatise that established the general principles of expression and the means of expressions in both humans and animals [4]. He also grouped various kinds of expressions into similar categories. The categorization is as follows:

- low spirits, anxiety, grief, dejection, despair
- joy, high spirits, love, tender feelings, devotion
- reflection, meditation, ill-temper, sulkiness, determination
- hatred, anger
- disdain, contempt, disgust, guilt, pride
- surprise, astonishment, fear, horror
- self-attention, shame, shyness, modesty

[1] Darwin has referred to *'Physiognomy'* as a subject that does not concern him [4].
[2] The references to the works of John Bulwer and Le Brun as can be found in Darwin's book [4].

Furthermore, Darwin also cataloged the facial deformations that occur for each of the above mentioned class of expressions. For example: *"the contraction of the muscles round the eyes when in grief"*, *"the firm closure of the mouth when in reflection"*, *"the depression of the corners of the mouth when in low spirits"*, etc [4][3].

Another important milestone in the study of facial expressions and human emotions is the work done by psychologist Paul Ekman and his colleagues since the 1970s. Their work is of significant importance and has a large influence on the development of modern day automatic facial expression recognizers. I have devoted a considerable part of section 3 and section 4 to give an introduction to their work and the impact it has had on present day systems.

Traditionally (as we have seen), facial expressions have been studied by clinical and social psychologists, medical practitioners, actors and artists. However in the last quarter of the 20th century, with the advances in the fields of robotics, computer graphics and computer vision, animators and computer scientists started showing interest in the study of facial expressions.

The first step towards the automatic recognition of facial expressions was taken in 1978 by Suwa et al. Suwa and his colleagues presented a system for analyzing facial expressions from a sequence of images (movie frames) by using twenty tracking points. Although this system was proposed in 1978, researchers did not pursue this line of study till the early 1990s. This can be clearly seen by reading the 1992 survey paper on the automatic recognition of faces and expressions by Samal and Iyengar [6]. The *'Facial Features and Expression Analysis'* section of the paper presents only four papers: two on automatic analysis of facial expressions and two on modeling of the facial expressions for animation. The paper also states that *"research in the analysis of facial expressions has not been actively pursued"* (page 74 from [6]). I think that the reason for this is as follows: The automatic recognition of facial expressions requires robust face detection and face tracking systems. These were research topics that were still being developed and worked upon in the 1980s. By the late 1980s and early 1990s, cheap computing power started becoming available. This led to the development of robust face detection and face tracking algorithms in the early 1990s. At the same time, Human-Computer Interaction and Affective Computing started gaining popularity. Researchers working on these fields realized that without automatic expression and emotion recognition systems[4], computers will remain cold and unreceptive to the users' emotional state. All of these factors led to a renewed interest in the development of automatic facial expression recognition systems.

Since the 1990s, (due to the above mentioned reasons) research on automatic facial expression recognition has become very active. Comprehensive and widely cited surveys by Pantic and Rothkrantz (2000) [7] and Fasel and Luttin (2003) [8] are available that perform an in-depth study of the published work from 1990 to 2001. Therefore, I will be mostly concentrating only on the published papers from 2001 onwards. Readers interested in the previous work can refer to the above mentioned surveys.

Till now we have seen a brief 'timeline view' of the important studies on expressions and expression recognition (right from the Aristotelian era till 2001). The remainder of the paper will be organized as follows: Section 2 mentions some of the applications of automatic facial expression recognition systems, section 3 gives the important techniques used to facial parameterization, section 4 gives a note on facial expressions and features, section 5 gives the characteristics of a good automatic face expressions recognition system, section 6 covers the various techniques used for face detection, tracking and feature extraction, section 7 gives a note on the various databases that have been used, section 8 gives a summary of the state of the art, section 9 gives a note on classifiers, section 10 gives an interesting note on the 6 prototypic expressions, section 11 mentions the challenges and future work and the paper concludes with section 12.

## 2. Applications

Automatic face expression recognition systems find applications in several interesting areas. With the recent advances in robotics, especially humanoid robots, the urgency in the requirement of a robust expression recognition system is evident. As robots begin to interact more and more with humans and start becoming a part of our living spaces and work spaces, they need to become more intelligent in terms of understanding the human's moods and emotions. Expression recognition systems will help in creating this intelligent visual interface between the man and the machine.

---

[3] It is important to note that Darwin's work is based on the work of Sir Charles Bell, one of his early contemporaries. In Darwin's own words, *"He (Sir Charles Bell) may with justice be said, not only to have laid the foundations of the subject as a branch of science, but to have built up a noble structure"* (page 2 from [4]). Readers interested in Sir Charles Bell's work can refer [5].
[4] Facial Expression recognition should not be confused with human Emotion Recognition. As Fasel and Luttin point out, *"Facial Expression recognition deals with the classification of facial motion and facial feature deformation into classes that are purely based on visual information"* whereas *"Emotion Recognition is an interpretation attempt and often demands understanding of a given situation, together with the availability of full contextual information"* (page 259 and 260 in [7]).

Humans communicate effectively and are responsive to each other's emotional states. Computers must also gain this ability. This is precisely what the Human-Computer Interaction research community is focusing on: namely, Affective Computing. Expression recognition plays a significant role in recognizing one's affect and in turn helps in building meaningful and responsive HCI interfaces. The interested reader can refer to Zeng et al.'s comprehensive survey [9] to get a complete picture on the recent advances in Affect-Recognition and its applications to HCI.

Apart from the two main applications, namely robotics and affect sensitive HCI, expression recognition systems find uses in a host of other domains like Telecommunications, Behavioral Science, Video Games, Animations, Psychiatry, Automobile Safety, Affect sensitive music juke boxes and televisions, Educational Software, etc.

Practical real-time applications have also been demonstrated. Bartlett et al. have successfully used their face expression recognition system to develop an animated character that mirrors the expressions of the user (called the *CU Animate*) [13]. They have also been successful in deployed the recognition system on Sony's *Aibo* Robot and ATR's *RoboVie* [13]. Another interesting application has been demonstrated by Anderson and McOwen, called the *'EmotiChat'* [18]. It consists of a chat-room application where users can log in and start chatting. The face expression recognition system is connected to this chat application and it automatically inserts emoticons based on the user's facial expressions.

As expression recognition systems become more real-time and robust, we will be seeing many other innovative applications and uses.

## 3. Facial Parameterization

The various facial behaviors and motions can be parameterized based on muscle actions. This set of parameters can then be used to represent the various facial expressions. Till date, there have been two important and successful attempts in the creation of these parameter sets:

1. The Facial Action Coding System (FACS) developed by Ekman and Friesen in 1977 [27] and
2. The Facial Animation parameters (FAPs) which are a part of the MPEG-4 Synthetic/Natural Hybrid Coding (SNHC) standard, 1998 [28].

Let us look at each of them in detail:

### 3.1. The Facial Action Coding System (FACS)

Prior to the compilation of the FACS in 1977, most of the facial behavior researchers were relying on the human observers who would observe the face of the subject and give their analysis. But such visual observations cannot be considered as an exact science since the observers may not be reliable and accurate. Ekman et al. questioned the validity of such observations by pointing out that the observer may be influenced by context [29]. They may give more prominence to the voice rather than the face and furthermore, the observations made may not be the same across cultures; different cultural groups may have different interpretations [29].

The limitations that the observers pose can be overcome by representing expressions and facial behaviors in terms of a fixed set of facial parameters. With such a framework in place, only these individual parameters have to be observed without considering the facial behavior as a whole. Even though, since the early 1920s researchers were trying to measure facial expressions and develop a parameterized system, no consensus had emerged and the efforts were very disparate [29]. To solve these problems, Ekman and Friesen developed the comprehensive FACS system which has since then become the de-facto standard.

Facial Action Coding is a muscle-based approach. It involves identifying the various facial muscles that individually or in groups cause changes in facial behaviors. These changes in the face and the underlying (one or more) muscles that caused these changes are called Action Units (AU). The FACS is made up of several such action units. For example:

- AU 1 is the action of raising the Inner Brow. It is caused by the *Frontalis* and *Pars Medialis* muscles,
- AU 2 is the action of raising the Outer Brow. It is caused by the *Frontalis* and *Pars Lateralis* muscles,

- AU 26 is the action of dropping the Jaw. It is caused by the *Masetter*, *Temporal* and *Internal Pterygoid* muscles,

and so on [29]. However not all of the AUs are caused by facial muscles. Some such examples are:

- AU 19 is the action of 'Tongue Out',
- AU 33 is the action of 'Cheek Blow',
- AU 66 is the action of 'Cross-Eye',

and so on [29]. The interested reader can refer to the FACS manuals [27] and [29] for the complete list of AUs.

AUs can be additive or non-additive. AUs are said to be additive if the appearance of each AU is independent and the AUs are said to be non-additive if they modify each other's appearance [30]. Having defined these, representation of facial expressions becomes an easy job. Each expression can be represented as a combination of one or more additive or non-additive AUs. For example 'fear' can be represented as a combination of AUs 1, 2 and 26 [30]. Figs. 1 and 2 show some examples of upper and lower face AUs and the facial movements that they produce when presented in combination.

| NEUTRAL | AU 1 | AU 2 | AU 4 | AU 5 |
|---|---|---|---|---|
| 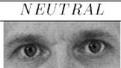 | 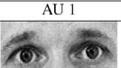 | 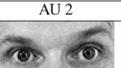 | 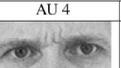 | 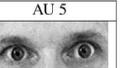 |
| Eyes, brow, and cheek are relaxed. | Inner portion of the brows is raised. | Outer portion of the brows is raised. | Brows lowered and drawn together | Upper eyelids are raised. |
| AU 6 | AU 7 | AU 1+2 | AU 1+4 | AU 4+5 |
| 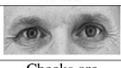 | 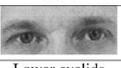 | 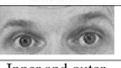 | 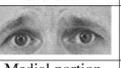 | 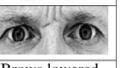 |
| Cheeks are raised. | Lower eyelids are raised. | Inner and outer portions of the brows are raised. | Medial portion of the brows is raised and pulled together. | Brows lowered and drawn together and upper eyelids are raised. |
| AU 1+2+4 | AU 1+2+5 | AU 1+6 | AU 6+7 | AU 1+2+5+6+7 |
| 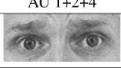 | 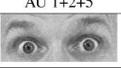 | 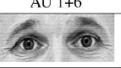 | 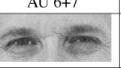 | 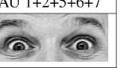 |
| Brows are pulled together and upward. | Brows and upper eyelids are raised. | Inner portion of brows and cheeks are raised. | Lower eyelids cheeks are raised. | Brows, eyelids, and cheeks are raised. |

Fig. 1: Some of the Upper Face AUs and their combinations. Figure reprinted from [10] [A].

| NEUTRAL | AU 9 | AU 10 | AU 12 | AU 20 |
|---|---|---|---|---|
| 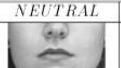 | 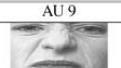 | 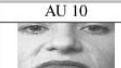 | 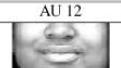 | 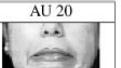 |
| Lips relaxed and closed. | The infraorbital triangle and center of the upper lip are pulled upwards. Nasal root wrinkling is present. | The infraorbital triangle is pushed upwards. Upper lip is raised. Causes angular bend in shape of upper lip. Nasal root wrinkle is absent. | Lip corners are pulled obliquely. | The lips and the lower portion of the nasolabial furrow are pulled pulled back laterally. The mouth is elongated. |
| AU15 | AU 17 | AU 25 | AU 26 | AU 27 |
| 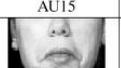 | 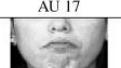 | 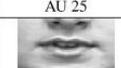 | 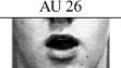 | 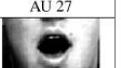 |
| The corners of the lips are pulled down. | The chin boss is pushed upwards. | Lips are relaxed and parted. | Lips are relaxed and parted; mandible is lowered. | Mouth stretched open and the mandible pulled downwards. |
| AU 23+24 | AU 9+17 | AU9+25 | AU9+17+23+24 | AU10+17 |
| 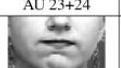 | 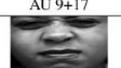 | 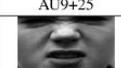 | 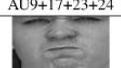 | 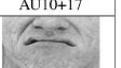 |
| Lips tightened, narrowed, and pressed together. | | | | |
| AU 10+25 | AU 10+15+17 | AU 12+25 | AU12+26 | AU 15+17 |
| 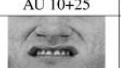 | 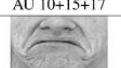 | 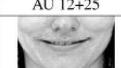 | 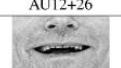 | 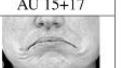 |
| AU 17+23+24 | AU 20+25 | | | |
| 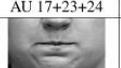 | 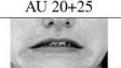 | | | |

Fig. 2: Some of the Lower Face AUs and their combinations. Figure reprinted from [10] [B].

There are other observational coding schemes that have been developed, many of which are variants of the FACS. They are: FAST (Facial Action Scoring Technique), EMFACS (Emotional Facial Action Coding System), MAX (Maximally Discriminative Facial Movement Coding System), EMG (Facial Electromyography), AFFEX (Affect Expressions by holistic Judgment), Mondic Phases, FACSAID (FACS Affect Interpretation Database) and Infant/Baby FACS. Readers can refer to Sayette et al. [38] for descriptions, comments and references on many of these systems. Also, the comparison of FACS with MAX, EMG, EMFACS and FACSAID has been given by Ekman and Rosenberg (pages 13-16 from [45]).

Automatic recognition of the AUs from the given image or video sequence is a challenging task. Recognizing the AUs will help in determining the expression and in turn the emotional state of the person. Tian et al. have developed the Automatic Face Analysis (AFA) system which can automatically recognize six upper face AUs and ten lower face AUs [10]. However real time applications may demand recognition of AUs from profile views too. Also, there are certain AUs like AU36t (tongue pushed forward under the upper lip) and AU29 (pushing jaw forward) that can be recognized only from profile images [20]. To address these issues, Pantic and Rothkrantz have worked on the automatic AU coding of profile images [15], [20].

Let us close this section with a discussion about the comprehensiveness of the FACS AU coding system. Harrigan et al. have discussed the difference between selectivity and comprehensiveness in the measurement of the different types of facial movement [81]. From their discussion, it is clear that comprehensive measurement and parameterization techniques are required to answer difficult questions like: Do people with different cultural and economic backgrounds show different variations of facial expressions when greeting each other? How does a person's facial expression change when he or she experiences a sudden change in heart rate? [81]. In order to test if the FACS parameterization system was comprehensive enough, Ekman and Friesen tried out all the permutations of AUs using their own faces. They were able to generate more than 7000 different combinations of facial muscle actions and concluded that the FACS system was comprehensive enough (page 22 from [81]).

### 3.2. The Facial Animation Parameters (FAPs)

In the 1990s and prior to that, the computer animation research community faced similar issues that the face expression recognition researchers faced in the pre-FACS days. There was no unifying standard and almost every animation system that was developed had its own defined set of parameters. As noted by Pandzic and Forchheimer, the efforts of the animation and graphics researchers were more focused on the facial movements that the parameters caused, rather than the efforts to choose the best set of parameters [31]. Such an approach made the systems unusable across domains. To address these issues and provide for a standardized facial control parameterization, the Moving Pictures Experts Group (MPEG) introduced the Facial Animation (FA) specifications in the MPEG-4 standard. Version 1 of the MPEG-4 standard (along with the FA specification) became the international standard in 1999.

In the last few years, face expression recognition researchers have started using the MPEG-4 metrics to model facial expressions [32]. The MPEG-4 standard supports facial animation by providing Facial Animation Parameters (FAPs). Cowie et al. indicate the relationship between the MPEG-4 FAPs and FACS AUs: *"MPEG-4 mainly focusing on facial expression synthesis and animation, defines the Facial Animation parameters (FAPs) that are strongly related to the Action Units (AUs), the core of the FACS"* (page 125 from [32]). To better understand this relationship between FAPs and AUs, I give a brief introduction to some of the MPEG-4 standards and terminologies that are relevant to face expression recognition. The explanation that follows in the next few paragraphs has been derived from [28], [31], [32], [33], [34] and [35] (readers interested in a detailed overview of the MPEG-4 Facial Animation technology can refer to the survey paper by Abrantes and Pereira [34]. For a complete in-depth understanding of the MPEG-4 standards, refer to [28] and [31]. Raouzaiou et al. give a detailed note on FAPs, their relation to FACS and the modeling of facial expressions using FAPs [35]).

The MPEG-4 defines a face model in its neutral state to have a specific set of properties like a) all face muscles are relaxed; b) eyelids are tangent to the iris; c) pupil is 1/3rd the diameter of the iris and so on. Key features like eye separation, iris diameter, etc are defined on this neutral face model [28].

The standard also defines 84 key feature points (FPs) on the neutral face [28]. The movement of the FPs is used to understand and recognize facial movements (expressions) and in turn also used to animate the faces. Fig. 3 shows the location of the 84 FPs on a neutral face as defined by the MPEG-4 standard.

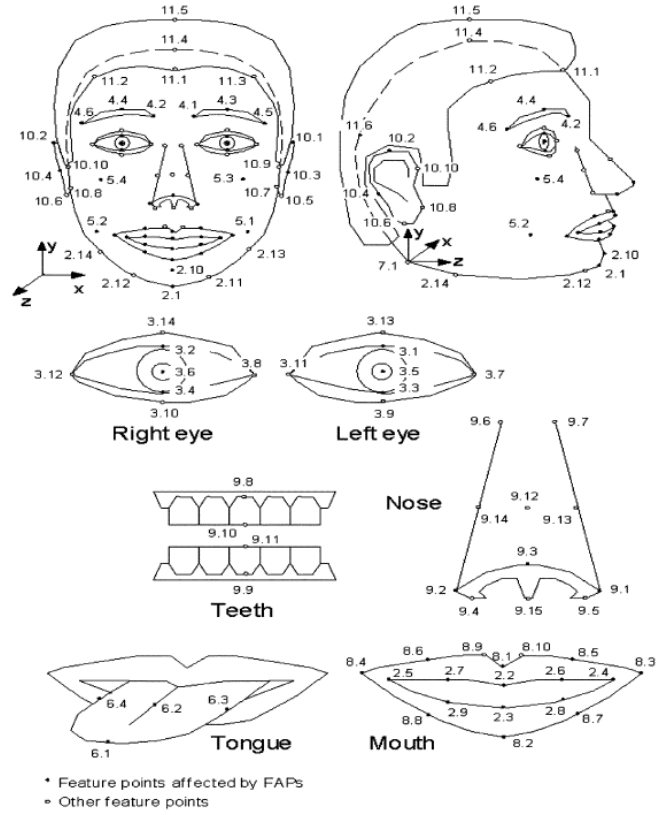

Fig. 3: The 84 Feature Points (FPs) defined on a neutral face. Figure reprinted from [28] [C].

The FAPs are a set of parameters that represent a complete set of facial actions along with head-motion, tongue, eye and mouth control (we can see that the FAPs like the AUs are closely related to muscle actions). In other words, each FAP is a facial action that deforms a face model in its neutral state. The FAP value indicates the magnitude of the FAP which in turn indicates the magnitude of the deformation that is caused on the neutral model, for example: a small smile versus a big smile. The MPEG-4 defines 68 FAPs [28]. Some of the FAPs and their description can be found in Table 1.

| FAP No. | FAP Name | FAP Description |
|---------|----------|-----------------|
| 3 | open_jaw | Vertical jaw displacement |
| 4 | lower_t_midlip | Vertical top middle inner lip displacement |
| 5 | raise_b_midlip | Vertical bottom middle inner lip displacement |
| 6 | stretch_l_cornerlip | Horizontal displacement of left inner lip corner |
| 7 | stretch_r_cornerlip | Horizontal displacement of right inner lip corner |

Table 1: Some of the FAPs and their descriptions [28]

In order to use the FAP values for any face model, the MPEG-4 standard defines normalization of the FAP values. This normalization is done using Facial Animation Parameter Units (FAPUs). The FAPUs are defined as a fraction of the distances between key facial features. Fig. 4 shows the face model in its neutral state along with the defined FAPUs. The FAP values are expressed in terms of these FAPUs. Such an arrangement allows for the facial movements described by a FAP to be adapted to any model of any size and shape.

The 68 FAPs are grouped into 10 FAP groups [28]. FAP group 1 contains two high level parameters: visemes (visual phonemes) and expressions. We are concerned with the expression parameter (the visemes are used for speech related studies). The MPEG-4 standard defines six primary facial expressions: joy, anger, sadness, fear, disgust and surprise (these are the 6 prototypical expressions that will be introduced in section 4). Although FAPs have been designed to allow the animation of faces with expressions, in recent years, the facial expression recognition community is working on the recognition of facial expressions and emotions using FAPs. For example the expression for 'sadness' can be expressed using FAPs as follows: *close_t_l_eyelid (FAP 19), close_t_r_eyelid (FAP 20), close_b_l_eyelid (FAP 22), raise_l_i_eyebrow (FAP 31), raise_r_i_eyebrow (FAP 32), raise_l_m_eyebrow (FAP 33), raise_r_m_eyebrow (FAP 34), raise_l_o_eyebrow (FAP 35), raise_r_o_eyebrow (FAP 36)* [35].

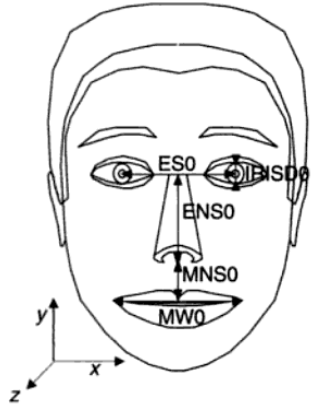

Fig. 4: The face model shown in its neutral state. The feature points used to define FAP units (FAPU) are also shown. The FAPUs are: ES0 (Eye Separation), IRISD0 (Iris Diameter), ENS0 (Eye-Nose Separation), MNS0 (Mouth-Nose Separation), MW0 (Mouth Width). Figure reprinted from [28][D].

As we can see, the MPEG-4 FAPs are closely related to the FACS AUs. Raouzaiou et al. [33] and Malatesta et al. [35] have worked on the mapping of AUs to FAPs. Some examples are shown in Table 2. Raouzaiou et al. have also derived models that will help in bringing together the disciplines of facial expression analysis and facial expression synthesis [35].

| Action Units (FACS AUs) | Facial Action Parameters (MPEG-4 FAPs) |
|---|---|
| AU1 (Inner Brow Raise) | raise_l_i_eyebrow + raise_r_i_eyebrow |
| AU2 (Outer Brow Raise) | raise_l_o_eyebrow + raise_r_o_eyebrow |
| AU9 (Nose Wrinkle) | lower_t_midlip + raise_nose + stretch_l_nose + stretch_r_nose |
| AU15 (Lip Corner Depressor) | lower_l_cornerlip + lower_r_cornerlip |

Table 2: Some of the AUs and their mapping to FAPs [35]

Recent works that are more relevant to the theme of this paper are the proposed automatic FAP extraction algorithms and their usage in the development of facial expression recognition systems. Let us first look at the proposed algorithms for the automatic extraction of FAPs. While working on an FAP based Automatic Speech Recognition system (ASR), Aleksic et al. wanted to automatically extract points from the outer lip (group 8), inner lip (group 2) and tongue (group 6). So, in order to automatically extract the FAPs, they introduced the Gradient Vector Flow (GVF) snake algorithm, Parabolic Template algorithm and a Combination algorithm [36]. While evaluating these algorithms, they found that the GVF snake algorithm, in comparison to the others, was more sensitive to random noise and reflections. Another algorithm that has been developed for automatic FAP extraction is the Active Contour algorithm by Pardas and Bonafonte [37]. These algorithms along with HMMs have been used successfully in the recognition of facial expressions [19], [37] (the system evaluation and results are presented in section 8).

In this section we have seen the two main facial parameterization techniques. Let us now take a look at some interesting points on emotions, facial expressions and facial features.

## 4. Emotions, Expressions and Features

One of the means of showing emotion is through changes in facial expressions. But are these facial expressions of emotion constant across cultures? For a long time, Anthropologists and Psychologists had been grappling with this question. Based on his theory of evolution, Darwin suggested that they were universal [40]. However the views were varied and there was no general consensus. In 1971, Ekman and Friesen conducted studies on subjects from western and eastern cultures and reported that the facial expressions of emotions were indeed constant across cultures [40]. The results of this study were well accepted. However, in 1994, Russell wrote a critique questioning the claims of universal recognition of emotion from facial expressions [42]. The same year, Ekman [43] and Izard [44] replied to this critique with strong evidences and gave a point to point refutation of Russell's claims. Since then, it has been considered as an established fact that the recognition of emotions from facial expressions is universal and constant across cultures.

Cross cultural studies have shown that, although the interpretation of expressions is universal across cultures, the expression of emotions though facial changes depend on social context [43], [44]. For example, when American and Japanese subjects were shown emotion eliciting videos, they showed similar facial expressions. However, in the presence of an authority, the Japanese viewers were much more reluctant to show their true emotions through changes in facial expressions [43]. In their cross cultural studies (1971), Ekman and Friesen used only six emotions, namely happiness, sadness, anger, surprise, disgust and fear. In their own words: *"the six emotions studied were those which had been found by more than one investigator to be discriminable within any one literate culture"* (page 126 from [40]). These six expressions have come to be known as the *'basic'*, *'prototypic'* or *'archetypal'* expressions. Since the early 1990s, researchers have been concentrating on developing automatic expression recognition systems that recognize these six basic expressions.

Apart from the six basic emotions, the human face is capable of displaying expressions for a variety of other emotions. In 2000, Parrott identified 136 emotional states that humans are capable of displaying and categorized them into separate classes and subclasses [39]. This categorization is shown in Table 3.

| Primary emotion | Secondary emotion | Tertiary emotions |
|---|---|---|
| Love | Affection | Adoration, affection, love, fondness, liking, attraction, caring, tenderness, compassion, sentimentality |
| | Lust | Arousal, desire, lust, passion, infatuation |
| | Longing | Longing |
| Joy | Cheerfulness | Amusement, bliss, cheerfulness, gaiety, glee, jolliness, joviality, joy, delight, enjoyment, gladness, happiness, jubilation, elation, satisfaction, ecstasy, euphoria |
| | Zest | Enthusiasm, zeal, zest, excitement, thrill, exhilaration |
| | Contentment | Contentment, pleasure |
| | Pride | Pride, triumph |
| | Optimism | Eagerness, hope, optimism |
| | Enthrallment | Enthrallment, rapture |
| | Relief | Relief |
| Surprise | Surprise | Amazement, surprise, astonishment |
| Anger | Irritation | Aggravation, irritation, agitation, annoyance, grouchiness, grumpiness |
| | Exasperation | Exasperation, frustration |
| | Rage | Anger, rage, outrage, fury, wrath, hostility, ferocity, bitterness, hate, loathing, scorn, spite, vengefulness, dislike, resentment |
| | Disgust | Disgust, revulsion, contempt |
| | Envy | Envy, jealousy |
| | Torment | Torment |
| Sadness | Suffering | Agony, suffering, hurt, anguish |
| | Sadness | Depression, despair, hopelessness, gloom, glumness, sadness, unhappiness, grief, sorrow, woe, misery, melancholy |
| | Disappointment | Dismay, disappointment, displeasure |
| | Shame | Guilt, shame, regret, remorse |
| | Neglect | Alienation, isolation, neglect, loneliness, rejection, homesickness, defeat, dejection, insecurity, embarrassment, humiliation |
| | Sympathy | Pity, sympathy |
| Fear | Horror | Alarm, shock, fear, fright, horror, terror, panic, hysteria, mortification |
| | Nervousness | Anxiety, nervousness, tenseness, uneasiness, apprehension, worry, distress, dread |

Table 3: The different emotion categories. Table reprinted from [41] [E] and the tree structure given by Parrott; pages 34 and 35 from [39].

In more recent years, there have been attempts at recognizing expressions other than the six basic ones. One of the techniques used to recognize non-basic expressions is by automatically recognizing the individual AUs which in turn helps in recognizing finer changes in expressions. An example of such a system is the Tian et al.'s AFA system that we saw in section 3.1 [10].

Most of the developed methods attempt to recognize the basic expressions and some attempts at recognizing non-basic expressions. However there have been very few attempts at recognizing the temporal dynamics of the face. Temporal dynamics refers to the timing and duration of facial activities. The important terms that are used in connection with temporal dynamics are: onset, apex and offset [45]. These are known as the temporal segments of an expression. Onset is the instance when the facial expression starts to show up, apex is the point when the expression is at its peak and offset is when the expression fades away (start-of-offset is when the expression starts to fade and the end-of-offset is when the expression completely fades out) [45]. Similarly onset-time is defined as the time taken from start to the peak, apex-time is defined as the total time at the peak and offset-time is defined as the total time from peak to the stop [45]. Pantic and Patras have reported successful recognition of facial AUs and their temporal segments. By doing so, they have been able to recognize a much larger range of expressions (apart from the prototypic ones) [46].

Till now we have seen the basic and non-basic expressions and some of the recent systems that have been developed to recognize expressions in each of the two categories. We have also seen the temporal segments associated with expressions. Now, let us now look at another very important type of expression classification: *'posed', 'modulated'* or *'deliberate'* versus *'spontaneous', 'unmodulated'* or *'genuine'*. Posed expressions are the artificial expressions that a subject will produce when he or she is asked to do so. This is usually the case when the subject is under normal test condition or under observation in a laboratory. In contrast, spontaneous expressions are the ones that people give out spontaneously. These are the expressions that we see on a day to day basis, while having conversations, watching movies etc. Since the early 1990s, till recent years, most of the researchers have focused on developing automatic face expression recognition systems for posed expressions only. This is due to the fact that posed expressions are easy to capture and recognize. Furthermore, it is very difficult to build a database that contains images and videos of subjects displaying spontaneous expressions. The specific problems faced and the approaches taken by the researchers will be covered in section 7.

Many psychologists have proved that posed expressions are different from spontaneous expressions, in terms of their appearance, timing and temporal characteristics. For an in-depth understanding of the differences between posed and spontaneous expressions, the interested reader can refer to chapters 9 through 12 in [45]. Furthermore, it also turns out that many of the posed expressions that researchers have used in their recognition systems are highly exaggerated. I have included fig. 5 as an example. It shows a subject (from one of the publicly available expression databases) displaying surprise, disgust and anger. It can be clearly seen that the expressions displayed by the subject are highly exaggerated and appear very artificial. In contrast, genuine expressions are usually subtle.

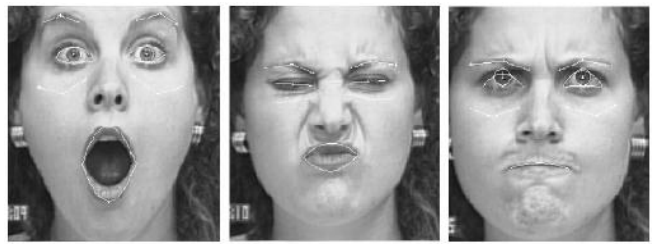

Fig. 5: Posed and exaggerated displays of Surprise, Disgust and Anger. Figure reprinted from [10][F].

The differences in posed and spontaneous expressions are quite apparent and this highlights the need for developing expression recognition systems that can recognize spontaneous rather than posed expressions. Looking forward, if the face expression recognition systems have to be deployed in real-time day to day applications, then they need to recognize the genuine expressions of humans. Although systems have been demonstrated that can recognize posed expressions with high recognition rates, they cannot be deployed in practical scenarios. Thus, in the last few years some of the researchers have started focusing on spontaneous expression recognition. Some of these systems will be discussed in section 8. The posed expression recognition systems are serving as the foundational work using which more practical spontaneous expression recognition systems are being built.

Let us now look at another class of expressions called 'microexpressions'. Microexpressions are the expressions that people display when they are trying to conceal their true emotions [86]. Unlike regular spontaneous expressions, microexpressions last only for 1/25th of a second. There has been no research done on the automatic recognition of microexpressions, mainly because of the fact that these expressions are displayed very quickly and are much more difficult to elicit and capture. Ekman has studied microexpressions in detail and has written extensively about the use of microexpressions to detect deception and lies (page 129 from [85]). Ekman also talks about 'squelched' expressions. These are the expressions that begin to show up but are immediately curtailed by the person by quickly changing his or her expression (page 131 from [85]). While microexpressions are complete in terms of their temporal parameters (onset-apex-offset), squelched expressions are not. But squelched expressions are found to last longer than microexpressions (page 131 from [85]).

Having discussed emotions and the associated facial expressions, let us now take a look at Facial Features. Facial features can be classified as being permanent or transient. Permanent features are the features like eyes, lips, brows and cheeks which remain permanently. Transient features include facial lines, brow wrinkles and deepened furrows that appear with changes in expression and disappear on a neutral face. Tian et al.'s AFA system uses recognition and tracking of permanent and transient features in order to automatically detect AUs [10].

Let us now look at the main parts of the face and the role they play in the recognition of expressions. As we can guess, the eyebrows and mouth are the parts of the face that carry the maximum amount of information related to the facial expression that is being displayed. This is shown to be true by Pardas and Bonafonte. Their work shows that surprise, joy and disgust have very high recognition rates (of 100%, 93.4% and 97.3% respectively) because they involve clear motion of the mouth and the eyebrows [37]. Another interesting result from their work shows that the mouth conveys more information than the eyebrows. The tests they conducted with only the mouth being visible gave a recognition accuracy of 78% whereas tests conducted with only the eyebrows visible gave a recognition accuracy of only 50%. Another occlusion related study by Bourel et al. has shown that sadness is mainly conveyed by the mouth [26]. Occlusion related studies are important because in real world scenarios, partial occlusion is not uncommon. Everyday items like sunglasses, shadows, scarves, facial hair, etc can lead to occlusions and the recognition system must be able to recognize expressions despite these occlusions [25]. Kotsia et al.'s study on the effect of occlusions on face expression recognizers shows that the occlusion of the mouth reduces the results by more than 50% [25]. This number perfectly matches with the results of Pardas and Bonafonte (that we have seen above). Thus we can say that, for expression recognition, the eyebrows and mouth are the most important parts of the face with the mouth being much more important than the eyebrows. Kotsia et al. also showed that occlusions on the left half or the right half of the face do not affect the performance [25]. This is because of the fact that the facial expressions are symmetric along the vertical plane that divides the face into left and right.

Another point to consider is the differences in facial features among individual subjects. Kanade et al. have written a note on this topic [66]. They discuss differences such as the texture of the skin and appearance of facial hair among adults and infants, eye opening in Europeans and Asians, etc (section 2.5 from [66]).

I will conclude this section with a debate that has been going on for a long time regarding the way the human brain recognizes faces (or in general, images). For a long time, psychologists and neuropsychologists have debated whether the recognition of the faces is by components or by a holistic process. While Biederman proposed the theory of recognition-by-components [83], Farah et al. suggested that recognition is a holistic process. Till date, there has been no agreement on the same. Thus, face expression recognition researchers have followed both approaches: holistic approaches like PCA and component based approaches like the use of Gabor Wavelets.

## 5. Characteristics of a Good System

We are now in a position where we can list down the features that a good face expression recognition system must possess:

- It must be fully automatic
- It must have the capability to work with video feeds as well as images
- It must work in real-time
- It must be able to recognize spontaneous expressions

- Along with the prototypic expressions, it must be able to recognize a whole range of other expressions too (probably by recognizing all the facial AUs)
- It must be robust against different lighting conditions
- It must be able to work moderately well even in the presence of occlusions
- It must be unobtrusive
- The images and video feeds do not have to be pre-processed
- It must be person independent
- It must work on people from different cultures and different skin colors. It must also be robust against age (in particular, recognize expressions of both infants, adults and the elderly)
- It must be invariant to facial hair, glasses, makeup etc.
- It must be able to work with videos and images of different resolutions
- It must be able to recognize expressions from frontal, profile and other intermediate angles

From the past surveys (and this one), we can see that different research groups have focused on addressing different aspects of the above mentioned points. For example, some have worked on recognizing spontaneous expression, some on recognizing expressions in the presence of occlusions; some have developed systems that are robust against lighting and resolution and so on. However going forward, researchers need to integrate all of these ideas together and build systems that can tend towards being ideal.

## 6. Face Detection, Tracking and Feature Extraction

Automatic face expression recognition systems are divided into three modules: 1) Face Tracking and Detection, 2) Feature Extraction and 3) Expression Classification. We will look at the first two modules in this section. A note on classifiers is presented in section 9.

The first step in facial expression analysis is to detect the face in the given image or video sequence. Locating the face within an image is termed as *face detection* or *face localization* whereas locating the face and tracking it across the different frames of a video sequence is termed as *face tracking*. Research in the fields of face detection and tracking has been very active and there is exhaustive literature available on the same. It is beyond the scope of this paper to introduce and survey all of the proposed methods. However this section will cover the face detection and tracking algorithms that have been used by face expression recognition researchers in the past few years (since the early 2000s).

One of the methods developed in the early 1990s to detect and track faces was the Kanade-Lucas-Tomasi tracker. The initial framework was proposed by Lucas and Kanade [56] and then developed further by Tomasi and Kanade [57]. In 1998, Kanade and Schneiderman developed a robust face detection algorithm using statistical methods [55], which has been used extensively since its proposal[5]. In 2004, Viola and Jones developed a method using the AdaBoost learning algorithm that was very fast and could rapidly detect frontal view faces. They were able to achieve excellent performance by using novel methods that could compute the features very quickly and then rapidly separate the background from the face [54]. Other popular detection and tracking methods were proposed by Sung and Poggio in 1998, Rowley et al. in 1998 and Roth et al. in 2000. For references and a comparative study of the above mentioned methods, interested readers can refer to [54] and [55].

A face tracker that is being used extensively by the face expression recognition researchers is the Piecewise Bezier Volume Deformation (PBVD) tracker, developed by Tao and Huang [47]. The tracker uses a generic 3D wireframe model of the face which is associated with 16 Bezier volumes. The wireframe model and the associated real-time face tracking are shown in fig. 6. It is interesting to note that the same PBVD model can also be used for the analysis of facial motion and the computer animation of faces.

Let us now look at face tracking using the 3-D Candide face model. The Candide face model is constructed using a set of triangles. It was originally developed by Rydfalk in 1987 for model based computer animation [48]. The face model was later modified by Stromberg, Welsh and Ahlberg and their models have come to be known as Candide-1, Candide-2 and Candide-3 respectively [59]. The 3 models are shown in Figs. 7a, 7b and 7c. Candide-3 is currently being used by most of the researchers. In recent years, apart from facial animation, the Candide face model has been used for face tracking as well [60]. It is one of the popular tracking techniques that are currently being used by face expression recognition researchers.

---

[5] The proposed detection method was considered to be *"influential and one that has stood the test of time"*. For this, Kanade and Schneiderman were awarded the 2008 Longuet-Higgins Prize at IEEE Conf. CVPR. (http://www.cmu.edu/qolt/News/2008%20Archive/kanade-longuet-higgin.html)

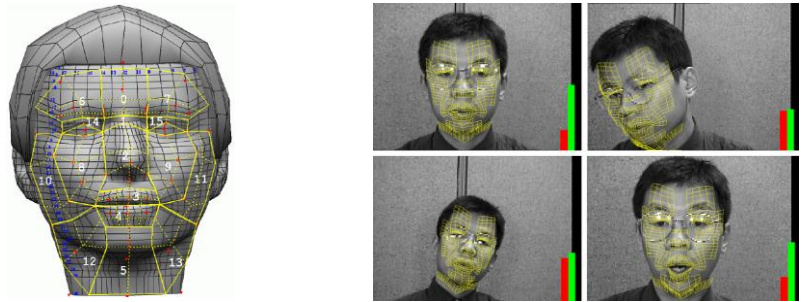

Fig. 6: The 16 Bezier volumes associated with a wireframe mesh and the real-time tracking that is achieved. Figure reprinted from [47][G].

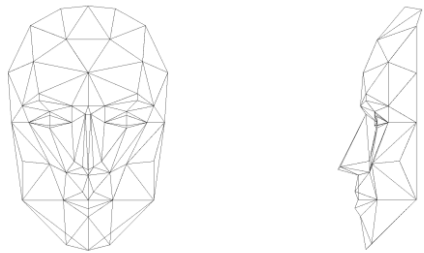

Fig. 7a: Candide - 1 face model. Figure reprinted from [47][H].

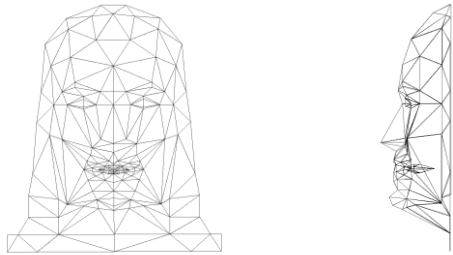

Fig. 7b: Candide - 2 face model. Figure reprinted from [47][I].

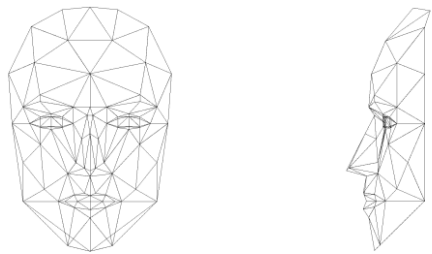

Fig. 7c: Candide - 3 face model. Figure reprinted from [47][J].

Another face tracker is the Ratio Template Tracker. The Ratio Template Algorithm was proposed by Sinha in 1996 for the cognitive robotics project at MIT [49]. Scassellati used it to develop a system that located the eyes by first detecting the face in real time [50]. In 2004, Anderson and McOwen modified the Ratio Template by incorporating the 'Golden Ratio' (or Divine Proportion) [61] into it. They called this modified tracker as the Spatial Ratio Template tracker [51]. This modified tracked was shown to work better under different illuminations. Anderson and McOwen suggest that this improvement is because of the incorporated 'Golden Ratio' which helps in describing the structure of the human face more accurately. The Spatial Ratio Template tracker (along with SVMs) has been successfully used in face expression recognition [18] (the evaluation is covered in section 8). The Ratio Template and the Spatial Ratio Template are shown in fig. 8a and their application to a face is shown in fig. 8b.

Another system that achieves fast and robust tracking is the PersonSpotter system [52]. It was developed in 1998 by Steffens et al. for real-time face recognition. Along with face recognition, the system also has modules for face detection and tracking. The tracking algorithm locates regions of interest which contain moving objects by forming difference images. Skin detectors and convex detectors are then applied to this region in order to detect and track the face. The PersonSpotter's tracking system has been demonstrated to be robust against considerable background motion. In the subsequent modules, the PersonSpotter system applies a model graph on the face and automatically suppressing the background. This is as shown in fig. 9.

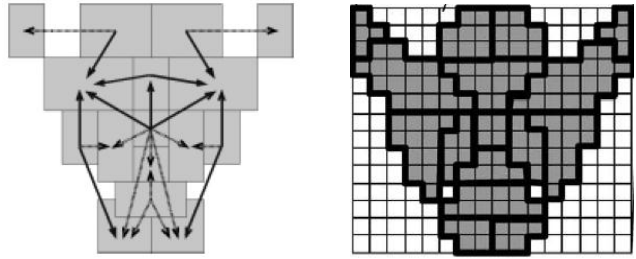

Fig. 8a: The image on the left is the Ratio Template (14 pixels by 16 pixels). The template is composed of 16 regions (gray boxes) and 23 relations (arrows). The image on the right is the Spatial Ratio Template. It is a modified version of the Ratio Template by incorporating the 'Golden Ratio'. Figures reprinted from [50] and [51] respectively[K].

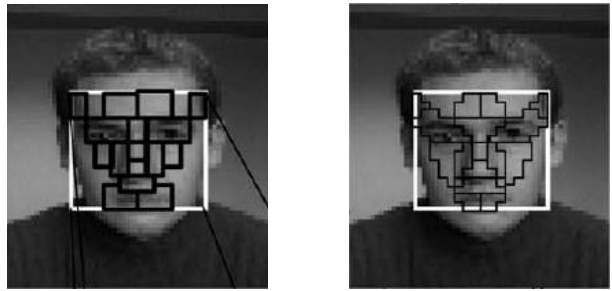

Fig. 8b: The image on the left shows the Ratio Template applied to a face. The image on the right shows the Spatial Ratio Template applied to a face. Figure reprinted from [51][L].

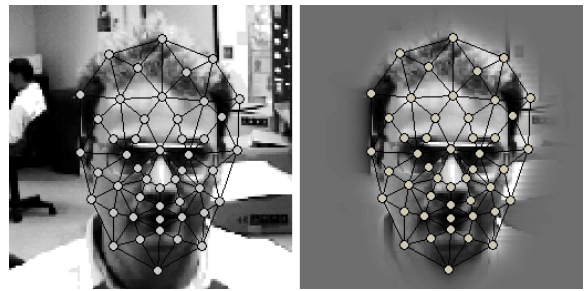

Fig. 9: PersonSpotter's model graph and background suppression. Figure reprinted from [52][M].

In 2000, Bourel et al., proposed a modified version of the Kanade-Lucas-Tomasi tracker [82], which they called as the EKLT tracker (Enhanced KLT tracker). They used the configurations and visual characteristics of the face to extend the KLT tracker which made the KLT tracker robust against temporary occlusions and allowed it to recover from the loss of points caused by tracking drifts.

Other popular face trackers (or in general, object trackers) are the ones based on Kalman Filters, extended Kalman Filters, Mean-Shifting and Particle Filtering. Out of these methods, particle filters (and adaptive particle filters) are more extensively used since they can deal successfully with noise, occlusion, clutter and a certain amount of uncertainty [62]. For a tutorial on particle filters, interested readers can refer to Arulampalam et al. [62]. In recent years, several improvements in particle filters have been proposed, some of which are mentioned here: Motion prediction has been used in adaptive particle filters to follow fast movements and deal with occlusions [63], mean shift has been incorporated into particle filters to deal with the degeneracy problem [64] and AdaBoost has been used with particle filters to allow for detection and tracking of multiple targets [65].

Other interesting approaches have been proposed. Instead of using a face tracker, Tian et al. have used multistate face models to track the permanent and transient features of the face separately [10]. The permanent features (lips, eyes, brow and cheeks) are tracked using a separate lip tracker [66], eye tracker [67] and brow and cheek tracker [56]. They used canny edge detection to detect the transient features like furrows.

## 7. Databases

One of the most important aspects of developing any new recognition or detection system is the choice of the database that will be used for testing the new system. If a common database is used by all the researchers, then testing the new system, comparing it with the other state of the art systems and benchmarking the performance becomes a very easy and straightforward job. However, building such a 'common' database that can satisfy the various requirements of the problem domain and become a standard for future research is a difficult and challenging task. With respect to face recognition, this problem is close to being solved with the development of the FERET Face Database which has become a de-facto standard for testing face recognition systems. However the problem of a standardized database for face expression recognition is still an open problem. But when compared to the databases that were available when the previous surveys [7], [8] were written, significant progress has been made in developing standardized expression databases.

When compared to face recognition, face expression recognition poses a very unique challenge in terms of building a standardized database. This challenge is due to the fact that expressions can be posed or spontaneous. As we have seen in the previous sections, there is a huge volume of literature in psychology that states that posed and spontaneous expressions are very different in their characteristics, temporal dynamics and timings. Thus, with the shifting focus of the research community from posed to spontaneous expression recognition, a standardized training and testing database is required that contains images and video sequences (at different resolutions) of people displaying spontaneous expressions under different conditions (lighting conditions, occlusions, head rotations, etc).

The work done by Sebe and colleagues [21] is one of the initial and important steps taken in the direction of creating a spontaneous expression database. To begin with, they listed the major problems that are associated with capturing spontaneous expressions. Their main observations were as follows:

- Different subjects express the same emotions at different intensities
- If the subject becomes aware that he or she is being photographed, their expression loses its authenticity
- Even if the subject is not aware of the recording, the laboratory conditions may not encourage the subject to display spontaneous expressions.

Sebe and colleagues consulted with fellow psychologists and came up with a method to capture spontaneous expressions. They developed a video kiosk where the subjects could watch emotion inducing videos. The facial expressions of were recorded by a hidden camera. Once the recording was done, the subjects were notified of the recording and asked for their permission to use the captured images and videos for research studies. 50% of the subjects gave consent. The subjects were then asked about the true emotions that they had felt. Their replies were documented against the recordings of the facial expressions [21].

From their studies, they found out that it was very difficult to induce a wide range of expressions among the subjects. In particular, fear and sadness were found to be difficult to induce. They also found that spontaneous facial expressions could be misleading. Strangely, some subjects had a facial expression that looked sad when they were actually feeing happy. As a side observation, they found that students and younger faculty were more ready in giving their consent to be included in the database whereas older professors were not [21].

Sebe et al.'s study helps us in understanding the problems encountered with capturing spontaneous expressions and the round-about mechanisms that have to be used in order to elicit and capture authentic facial expressions.

Let us now look at some of the popular expression databases that are publicly and freely available. There are many databases available and covering all of them will not be possible. Thus, I will be covering only those databases that have mostly been used in the past few years. To begin with, let us look at the Cohn-Kanade database also known as the CMU-Pittsburg AU coded database [66], [67]. This is a fairly extensive database (figures and facts are presented in table 4) and has been widely used by the face expression recognition community. However this database has only posed expressions… 

| Database | Sample Details | Expression Elicitation and Data Recording Methods | Available Descriptions | Additional Notes | Reference |
|---|---|---|---|---|---|
| Cohn-Kanade Database (also known as CMU-Pittsburg database) [66], [67]. | • 500 image sequences from 100 subjects<br>• Age: 18 to 30 years<br>• Gender: 65% female<br>• Ethnicity: 15% African - Americans and 3% of Asians and Latinos | This DB contains only posed expressions. "The subjects were instructed to perform a series of 23 facial displays that included single action units (e.g. AU 12, or lip corners pulled obliquely) and action unit combinations (e.g. AU 1+2, or inner and outer brows raised). Each begins from a neutral or near-ly neutral face. For each, an experimenter described and modeled the target display. Six were based on descriptions of prototypic emotions (i.e., joy, surprise, anger, fear, disgust, and sadness). These six tasks and mouth opening in the absence of other action units were annotated by certified FACS coders." | "Annotation of FACS Action Units and emotion-specified expressions" | "Images taken using 2 cameras: one directly in front of the subject and the other positioned 30 degrees to the subject's right. But the DB contains only the images taken from the frontal camera." | Information presented here has been quoted from [67]. |
| MMI Facial Expression Database [69], [74] | • 52 different subjects of both sexes<br>• Gender: 48% female<br>• Age: 19 to 62 years<br>• Ethnicity: European, Asian, or South American<br>• Background: Natural lighting and variable backgrounds (for some samples) | This DB contains posed and spontaneous expressions. "The subjects were asked to display 79 series of expressions that included either a single AU (e.g., AU2) or a combination of a minimal number of AUs (e.g., AU8 cannot be displayed without AU25) or a prototypic combination of AUs (such as in expressions of emotion). Also, a short neutral state is available at the beginning and at the end of each expression."<br><br>For natural expressions: "Children interacted with a comedian. Adults watching emotion inducing videos" | "Action Units, metadata (data format, facial view, shown AU, shown emotion, gender, age), analysis of AU temporal activation patterns" | "The emotions were determined using an expert annotator".<br><br>A highlight of this DB is that it contains both frontal and profile view images. | Information presented here has been quoted from [75] |
| Spontaneous Expressions Database [21] | • 28 subjects | This DB contains spontaneous expressions. The subjects were asked to watch emotion inducing videos in a custom built video kiosk. Their expressions were recorded using hidden cameras. Then, the subjects were informed about the recording and were asked for their consent. Out of 60, 28 gave consent. | The database is self labeled. After watching the videos, the subjects recorded the emotions that they felt. | The researchers found that it is very difficult to induce all the emotions in all of the subjects. Joy, surprise and disgust were the most easy whereas sadness and fear were the most difficult. | Information presented here has been quoted from [21] |
| The AR Face Database [78], [79] | • 126 people<br>• Gender: 70 men and 56 women<br>• Over 4000 color images are available | This DB contains only posed expressions. "No restrictions on wear (clothes, glasses, etc.), make-up, hair style, etc. were imposed to participants. Each person participated in two sessions, separated by two weeks time. The same pictures were taken in both sessions." | None | This database has frontal-faces with different expressions, illumination conditions and occlusions (scarf and sunglasses). | Information presented here has been quoted from [78] |
| CMU Pose, Illumination, Expression (PIE) Database [80] | • 41,368 images of 68 people<br>• 4 different expressions. | This DB contains only posed expressions. | None | This database provides facial images for 13 different poses 43 different illumination conditions. | Information presented here has been quoted from [80] |
| The Japanese Female Facial Expression (JAFFEE) Database [76] | • 219 images of 7 facial expressions (6 basic facial expressions + 1 neutral)<br>• 10 Japanese female models. | This DB contains only posed expressions. The photos have been taken under strict controlled conditions of similar lighting and with the hair tied away from the face. | "Each image has been rated on 6 emotion adjectives by 92 Japanese subjects" | All the expressions are multiple AU expressions. | Information presented here has been quoted from [76] and [77]. |

Table 4: A summary of some of the facial expression databases that have been used in the past few years.

*continued from pg. 14*… Looking forward, although this database may be used for comparison and benchmarking against previously developed systems, it will be not be suitable to use it for spontaneous expression recognition. Some of the other databases that contain only posed expressions are the AR face database [78], [79], the CMU Pose, Illumination and Expression (PIE) database [80] and the Japanese Female Facial Expression database (JAFFE) [76]. Table 4 gives the details of each of these databases.

A major step towards creating a standardized database that can meet the needs of the present day research community is the creation of the MMI Facial Expression database by Pantic and colleagues [69], [74]. The MMI facial expression database contains both posed and spontaneous expressions. It also contains profile view data. The highlight of this database is that it is web-based, fully searchable and downloadable for free. The database went online on February 2009 [75]. For specific numerical details please refer to table 4.

There is also a database created by Ekman and Hager. It is called the Ekman-Hager database or the Ekman-Hager Facial Action Exemplars [70]. Some other researchers have created their own databases (for example, Donato et al. [71], Chen-Huang [73]). Another important database is Ekman's datasets [72] (but not available for free). It contains various data from cross-cultural studies and recent neuropsychological studies, expressive and neutral faces of Caucasians and Japanese, expressive faces from some of the stone-age tribes, and data that demonstrate the difference between true enjoyment and social smiles [72]. Most of the research has been focused on 2D face expression recognition. But 3D face expression recognition has been mostly ignored. To address this issue, a 3D facial expression database has also been developed [68]. It contains 3D range data from a 3DMD digitizer. Apart from the above mentioned databases, there are many other databases like the RU-FACS-1, USC-IMSC, UA-UIUC, QU, PICS and others. A comparative study of some of these databases with the MMI database is given in [69].

Till now we have focused on the issues related to capturing data and the work that has been done on the same. However there is one more unaddressed issue. Once the data has been captured, it needs to be labeled and augmented with helpful metadata. Traditionally the expressions have been labeled by the help of expert observers or AU coders or with the help of the subjects themselves (by asking them what emotion they felt). However such labeling is time consuming and requires expertise on the part of the observer; for self observations, the subjects may need to be trained [11]. Cohen et al. observed that although labeled data is available in fewer quantities, there is a huge volume of unlabeled data that is available. So they used Bayesian network classifiers like Naïve Bayes (NB), Tree Augmented Naïve Bayes (TAN) and Stochastic Structure Search (SSS) for semi-supervised learning with could work with some amount of labeled data and large amounts of unlabeled data [11].

I will now give a short note on the problems that specific researchers have faced with respect to the use of available databases. Cohen et al. reported that they could not make use of the Cohn-Kanade database to train and test a Multi-Level HMM classifier because each video sequence ends in the peak of the facial expression, i.e. each sequence was incomplete in terms of its temporal pattern [12]. Thus, looking forward database creators must note that the temporal patterns of the video captures must be complete. That is, all the video sequences must have the complete pattern of onset, apex and offset captured in the recordings. Another specific problem was reported by Wang and Yin in their face expression recognition studies using topographic modeling [23]. They note that topographic modeling is a pixel based approach and is therefore not robust against illumination changes. But in order to conduct illumination related studies, they were unable to find a database that had expressive faces against various illuminations. Another example is the occlusion related studies conducted by Kotsia et al. [25]. They could not find any popularly used database that had expressive faces with occlusions. So they had to preprocess the Cohn-Kanade and JAFFE database and graphically add occlusions.

From the above discussion it is quite apparent that the creation of a database that will serve everyone's needs is a very difficult job. However, there have been new databases created that contain spontaneous expressions, frontal and profile view data, 3D data, data under varying conditions of occlusion, lighting, etc and are publicly and freely available. This is a very important factor that will have a positive impact on the future research in this area.

## 8. State of the Art

Since it is almost impossible to cover all of the published work, I have selected 19 papers that I felt were important and very different from each other. Many of these 19 papers have already been introduced in the previous sections. I will not get into the verbose descriptions of these papers; rather table 5 has been created that will give a summary of all the surveyed papers. The papers are presented in chronological order starting from 2001.

| Reference | Feature Extraction | Classifier | Database | Sample size | Performance | Important Points |
|---|---|---|---|---|---|---|
| Tian et al., 2001 [10] | Permanent features: Optical Flow, Gabor Wavelets and Multi-state Models. Transient features: Canny edge detection | 2 ANNs, one for upper face and one for lower face | Cohn-Kanade and Ekman-Hager Facial Action Exemplars | Upper Face: 50 sample sequences from 14 subjects performing 7 AUs Lower Face: 63 sample sequences from 32 subjects performing 11 AUs | Recognition of Upper Face AUs: 96.4% Recognition of Lower Face AUs: 96.7% Average of 93.3% when generalized to independent databases | Recognizes posed expressions. Real-time system. Automatic Face detection. Head motion is handled Invariant to scaling. Uses facial feature tracker to reduce processing time. |
| Bourel et al., 2001 [26] | Local spatio-temporal vectors obtained from the EKLT tracker | Modular classifier with data fusion. Local classifiers are rank weighted kNN classifiers. Combination is a sum scheme. | Cohn-Kanade | 30 subjects 25 sequences for 4 expressions (total of 100 video sequences) | Refer fig. 10 | Deals with recognizing facial expressions in the presence of occlusions. Proposes use of modular classifiers instead of monolithic classifiers. Classification is done locally and then the classifier outputs are fused. |
| Pardas and Bonafonte, 2002 [37] | MPEG-4 FAPs extracted using an improved Active Contour algorithm and motion estimation | HMM | Cohn-Kanade | Used the whole DB | Overall efficiency of 84% (across 6 prototypic expressions) Experiments with: joy, surprise and anger: 98%, joy, surprise and sadness: 95% | Automatic extraction of MPEG-4 FAPs Proves that FAPs convey the necessary information that is required to extract the emotions. |
| Cohen et al., 2003 [12] | A vector of extracted Motion Units (MUs) using PBVD tracker | NB, TAN, SSS, HMM, ML-HMM | Cohn-Kanade and own DB | Cohn-Kanade: 53 subjects Own DB: 5 subjects | Refer table 6 | Real-time system. Emotion classification from video. Suggests use of HMMs to automatically segment a video into different expression segments. |
| Cohen et al., 2003 [11] | A vector of extracted Motion Units (MUs) using PBVD tracker | NB, TAN and SSS | Cohn-Kanade and Chen-Huang | Refer table 7 | Refer table 8 | Real-time system. Uses semi-supervised learning to work with some labeled data and large amount of unlabeled data. |
| Bartlett et al., 2003 [13] | Gabor Wavelets | SVM, AdaSVM (SVM with AdaBoost). | Cohn-Kanade | 313 sequences from 90 subjects. First and last frame used as training images | SVM with Linear Kernel: Automatic face detection: 84.8%; Manual alignment: 85.3%. SVM with RBF Kernel: Automatic face detection: 87.5%; Manual alignment: 87.6% | Fully automatic system. Real-time recognition at high level of accuracy Successfully deployed on Sony's Aibo pet robot, ATR's RoboVie and CU Animator |
| Michel and Kaliouby, 2003 [14] | Vector of feature displacements (Euclidean distance between neutral and peak) | SVM | Cohn-Kanade | For each basic emotion, 10 examples were used for training and 15 examples were used for testing. | With RBF Kernel: 87.9%. Person independent: 71.8% Person dependent (train and test data supplied by expert): 87.5% Person dependent (train and test data supplied by 6 users during ad-hoc interaction): 60.7% | Real-time system. Does not require any preprocessing. |

Table 5: A summary of some of the posed and spontaneous expression recognition systems (since 2001).

| Reference | Feature Extraction | Classifier | Database | Sample size | Performance | Important Points |
|---|---|---|---|---|---|---|
| Pantic and Rothkrantz, 2004 [15] | Frontal and Profile facial points | Rule based classifier | MMI | 25 subjects | 86% accuracy | Recognizes facial expressions in frontal and profile views. Proposed a way to do automatic AU coding in profile images. Not Real-time. |
| Buciu and Pitas, 2004 [16] | Image representation using Non negative Matrix Factorization (NMF) and Local Non negative Matrix factorization (LNMF). | Nearest neighbor classifier using CSM and MCC | Cohn-Kanade and JAFFE | Cohn-Kanade: 164 samples JAFFE: 150 samples | Cohn-Kanade: LNMF with MCC gave the highest accuracy of 81.4% JAFFE: Only 55% to 68% for all 3 methods | PCA was also performed for comparison purpose. LNMF outperformed both PCA and NMF whereas NMF produces the poorest performance. CSM classifier is more reliable than MCC and gives better recognition. |
| Pantic and Patras, 2005 [46] | Tracking a set of 20 facial fiducial points | Temporal Rules | Cohn-Kanade and MMI | Cohn-Kanade: 90 images MMI: 45 images | Overall an average recognition of 90% | Recognizes 27 AUs. Invariant to occlusions like glasses and facial hair. Shown to give a better performance than the AFA system |
| Zheng et al., 2006 [17] | 34 landmark points converted into a Labeled Graph (LG) using Gabor wavelet transform. Then a semantic expression vector built for each training face. KCCA used to learn the correlation between LG vector and semantic vector. | The correlation that is learnt is used to estimate semantic expression vector which is then used for classification | JAFFE and Ekman's Pictures of Affect | JAFFE: 183 images Ekman's: 96 images Neutral expressions were not chosen from either database | Using Semantic Info: On JAFFE DB: with Leave one image out (LOIO) cross validation: 85.79%, with Leave one subject out (LOSO) cross validation: 74.32%, On Ekman's DB: 81.25% Using Class Label Info: On JAFFE DB: with LOIO: 98.36%, with LOSO: 77.05%, On Ekman's DB: 78.13% | Used KCCA to recognize facial expressions The singularity problem of the Gram matrix has been tackled using an improved KCCA algorithm. |
| Anderson and McOwen, 2006 [18] | Motion signatures obtained by tracking using spatial ratio template tracker and performing optical flow on the face using multi-channel gradient model (MCGM) | SVM and MLP | CMU-Pittsburg AU coded DB and a non-expressive DB | CMU: 253 samples of 6 basic expressions. But these had to be preprocessed by reducing frame rate and scale Non-expressive: 10 subjects, 4800 frames long | Motion averaging using: co-articulation regions: 63.64%, 7x7 blocks: 77.92%, ratio template algorithms, with MLP: 81.82%, with SVM: 80.52% | Fully automated, multistage system. Real-time system. Able to operate efficiently in cluttered scenes. Used motion-averaging to condense the data that is fed to the classifier. |
| Aleksic and Katsaggelos, 2006 [19] | MPEG-4 FAPs, outer lip (group 8) and eyebrow (group 4) followed by PCA to reduce dimensionality | HMM and MS-HMM | Cohn-Kanade | 284 recordings of 90 subjects | Using HMM: Only eyebrow FAPs: 58.8%, Only outer lip FAPs: 87.32%, Joint FAPs: 88.73% After assigning stream weights and then using a MS-HMM: 93.66% with outer lip having more weight than eyebrows. | Showed that performance improvement is possible by using MS-HMMs and proposed a way to assign stream weights. Used PCA to reduce the dimensionality of the features before giving it to the HMM. |
| Pantic and Patras, 2006 [20] | Mid level parameters generated by tracking 15 facial points using particle filtering | Rule based classifier | MMI | 1500 samples of both static and profile views (single and multiple AU activations) | 86.6% on 96 test profile sequences | Automatic segmentation of input video into facial expressions. Recognition of temporal segments of 27 AUs occurring alone or in combination. Automatic recognition of AUs from profile images. |

Table 5 *(Continued)*: A summary of some of the posed and spontaneous expression recognition systems (since 2004).

| Reference | Feature Extraction | Classifier | Database | Sample size | Performance | Important Points |
|-----------|--------------------|------------|----------|-------------|-------------|------------------|
| Sebe et al., 2007 [21] | MUs generated from the PBVD tracker | Bayesian nets, SVMs and Decision Trees. Used voting algorithms like bagging and Boosting to improve results. | Created spontaneous emotions database. Also used Cohn-Kanade | Created DB: 28 subjects showing mostly neutral, joy, surprise and delight. Cohn-Kanade: 53 subjects | Using many different classifiers: Cohn-Kanade: 72.46% to 93.06%, Created DB: 86.77% to 95.57%. Using kNN with k = 3, best result of 93.57% | Recognizes spontaneous expressions Created an authentic DB where subjects are showing their natural facial expressions Evaluated several Machine Learning algorithms |
| Kotsia and Pitas, 2007 [22] | Geometric displacement of Candide nodes | Multiclass SVM: For expression recognition: used six-class SVM, one for each expression. For AU recognition: used one-class SVMs, one for each of the 8 chosen AUs used. | Cohn-Kanade | Whole DB | 99.7% for facial expression recognition 95.1% for facial expression recognition based on AU detection | Recognizes either the six basic facial expressions or a set of chosen AUs. Very high recognition rates have been shown |
| Wang and Yin, 2007 [23] | Topographic context (TC) expression descriptors | QDC, LDA, SVC and NB | Cohn-Kanade and MMI | Cohn-Kanade: 53 subjects, 4 images per subject for each expression. Total of 864 images MMI: 5 subjects, 6 images per subject for each expression. Total of 180 images | Person dependent tests: on MMI: with QDC: 92.78%, with LDA: 93.33%, with NB: 85.56%, on Cohn-Kanade: with QDC: 82.52%, with LDA: 87.27%, with NB: 93.29%. Person independent tests: on Cohn-Kanade: with QDC: 81.96%, with LDA: 82.68%, with NB: 76.12%, with SVC: 77.68% | Proposed a topographic modeling approach in which the gray scale image is treated as a 3D surface. Analyzed the robustness against the distortion of detected face region and the different intensities of facial expressions. |
| Dornaika and Davoine, 2008 [24] | Candide face model used to track features. | First head pose is determined using Online Appearance Models and then expressions are recognized using a stochastic approach | Created own data | Used several video sequences. Also created a challenge 1600 frame test video, where subjects were allowed to display any expression in any order for any duration | Results have been spread across different graphs and charts. Interested readers can refer [24] to view the same. | Proposes a framework for simultaneous face tracking and expression recognition. 2 AR models per expression gave better mouth tracking and in turn better performance. The video sequences contained posed expressions. |
| Kotsia et al., 2008 [25] | 3 approaches: Gabor features, DNMF algorithm and by Geometric displacement vectors extracted using Candide tracker | Multiclass SVM and MLP | Cohn-Kanade and JAFFE | | Using JAFFE: with Gabor: 88.1%, with DNMF: 85.2% Using Cohn-Kanade: with Gabor: 91.6%, with DNMF: 86.7%, with SVM: 91.4% | Developed a system to recognize expressions in-spite of occlusions. Discusses the effect of occlusion on the 6 prototypic facial expressions. |

Table 5 *(Continued)*: A summary of some of the posed and spontaneous expression recognition systems (since 2007).

**ANN:** Artificial Neural Network, **kNN:** k-Nearest Neighbor, **HMM:** Hidden Markov Model, **NB:** Naïve Bayes, **TAN:** Tree Augmented Naïve Bayes, **SSS:** Stochastic Structure Search, **ML-HMM:** Multi-Level HMM, **SVM:** Support Vector Machine, **CSM:** Cosine Similarity Measure, **MCC:** Maximum Correlation Classifier, **KCCA:** Kernel Canonical Correlation Analysis, **MLP:** Multi-Layer perceptron, **MS-HMM:** Multi-Stream HMM, **QDC:** Quadratic Discriminant Classifier, **LDA:** Linear Discriminant Classifier, **SVC:** Support Vector Classifier.

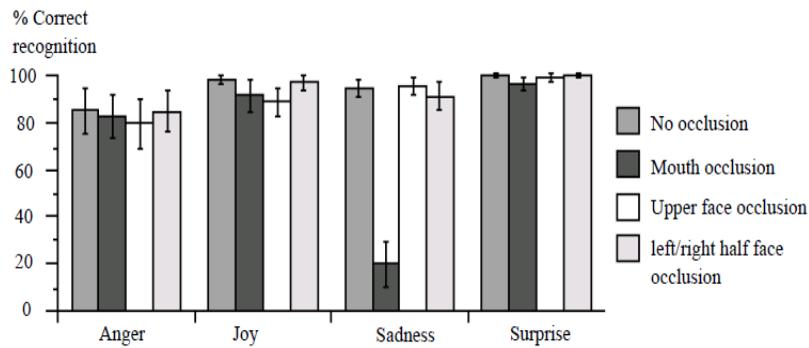

Fig. 10 *(referred from table 5)*: Bourel et al.'s system's performance. Figure reprinted from [26][N].

|  | **Cohn-Kanade DB** | **Own DB** |
|---|---|---|
| Person Dependent Tests | Insufficient data to perform person dependent tests | NB (Gaussian): 79.36% |
|  |  | NB (Cauchy): 80.05% |
|  |  | TAN: 83.31% |
|  |  | HMM: 78.49% |
|  |  | ML-HMM: 82.46% |
| Person Independent Tests | NB (Gaussian): 67.03% | NB (Gaussian): 60.23% |
|  | NB (Cauchy): 68.14% | NB (Cauchy): 64.77% |
|  | TAN: 73.22% | TAN: 66.53% |
|  | Insufficient data to conduct tests using HMM and ML-HMM | HMM: 55.71% |
|  |  | ML-HMM: 58.63% |

Table 6 *(referred from table 5)*: Cohen et al.'s system's performance [12]

|  | **Cohn-Kanade DB** | **Chen-Huang DB** |
|---|---|---|
| Labeled and Unlabeled data | 200 labeled, 2980 unlabeled for training and 1000 for testing | 300 labeled, 11982 unlabeled for training and 3555 for testing |
| Only Labeled data | 53 subjects displaying 4 to 6 expressions. 8 frames per expression sequence | 5 subjects displaying 6 expressions. 60 frames per expression sequence |

Table 7 *(referred from table 5)*: Cohen et al.'s sample size [11]

|  | **Cohn-Kanade DB** | **Chen-Huang DB** |
|---|---|---|
| Labeled and Unlabeled data | NB classifier: 69.10% | NB classifier: 58.54% |
|  | TAN classifier: 69.30% | TAN classifier: 62.87% |
|  | SSS classifier: 74.80% | SSS classifier: 74.99% |
| Only Labeled data | NB classifier: 77.70% | NB classifier: 71.78% |
|  | TAN classifier: 80.40% | TAN classifier: 80.31% |
|  | SSS classifier: 81.80% | SSS classifier: 83.62% |

Table 8 *(referred from table 5)*: Cohen et al.'s system performance [11]

## 9. Classifiers

The final stage of any face expression recognition system is the classification module (after the face detection and feature extraction modules). A lot of recent work has been done on the study and evaluation of the different classifiers.

Cohen et al. have studied static classifiers like the Naïve Bayes (NB), Tree Augmented Naïve Bayes (TAN), Stochastic Structure Search (SSS) and dynamic classifiers like Single Hidden Markov Models (HMM) and Multi-Level Hidden Markov Models (ML-HMM) [11], [12]. In the next few paragraphs, I will cover some of the important results of their study.

While studying NB classifiers, Cohen et al. experimented with the use of Gaussian distribution and Cauchy distribution as the model distribution of the NB classifier. They found that using the Cauchy distribution as the model gives better results [12]. Let us look at NB classifiers in more detail. We know about the independence assumption of NB classifiers. Although this independence assumption is not true in many of the real-world scenarios, NB classifiers are known to work surprisingly well. For example, if the word 'Bill' appears in an email, the probability of 'Clinton' or 'Gates' appearing become higher. This is a clear violation of the independence assumption. However NB classifiers have been used very successfully in classifying email as spam and non-spam. When it comes to face expression recognition, Cohen et al. suggest that similar independence assumption problems exist. This is so because there is a high degree of correlation between the display of emotions and facial motion [12]. They then studied the TAN classifier and found it to be better than NB. As a generalized thumb rule, they suggest the use of a NB classifier when data is insufficient since the TAN's learnt structure becomes unreliable and the use of TAN when sufficient data is available [12].

Cohen et al. have also suggested the scenarios where static classifiers can be used and the scenarios where dynamic classifiers can be used [12]. They report that dynamic classifiers are sensitive to changes in the temporal patterns and the appearance of expressions. So they suggest the use of dynamic classifiers when performing person dependent tests and the use of static classifiers when performing person independent tests. But there are other notable differences too: static classifiers are easy to implement and train when compared to dynamic classifiers. But on the flip side, static classifiers give poor performance when given expressive faces that are not at their apex [12].

As we saw in section 7, Cohen et al. used NB, TAN and SSS while working on semi-supervised learning to address the issues of labeled and unlabeled databases [11]. They found that NB and TAN performed well with training data that had been labeled. However they performed poorly when unlabeled data was added to the training set. So, they introduced the SSS algorithm and which could outperform NB and TAN when presented with unlabeled data [11].

Coming to dynamic classifiers, HMMs have traditionally been used to classify expressions. Cohen et al. suggested an alternative use of HMMs by using it to automatically segment arbitrarily long video sequences into different expressions [12].

Bartlett et al. used AdaBoost to speed up the process of feature selection. Improved classification performance was obtained by training the SVMs using this representation [13].

Bourel et al. proposed the localized representation and classification of features followed by the application of data fusion [26]. They propose the use of several modular classifiers instead of one monolithic classifier since the failure of one of the modular classifiers does not necessarily affect the classification. Also, the system becomes robust against occlusions; even if one part of the face is not visible then only that particular classifier fails whereas the other local classifiers still function properly [26]. Another reason stated was that new classification modules can be added easily if and when required. They used a rank weighted kNN classifier for each local classifier and the combination was implemented by a simple summation of classifier outputs [26].

Anderson and McOwen suggested the use of motion averaging over specified regions of the face to condense the data into an efficient form [18]. They reported that the classifiers worked better when fed with condensed data rather than the whole uncondensed data [18]. Next they studied MLPs and SVMs and found that both were giving almost similar performance. In order to choose among the two, they performed statistical studies and decided to go with the use of SVMs. A major factor affecting this decision was the fact that the SVMs had a much lower False Acceptance Rate (FAR) when compared to MLPs.

In 2007, Sebe et al. evaluated several machine leaning algorithms like Bayesian Nets, SVMs and Decision Trees [21]. They also used voting algorithms like bagging and boosting to improve the classification results. It is interesting to note that they found NB and kNN to be stable algorithms whose performance did not improve significantly with the usage of the voting algorithms. In total, they published evaluation results for 14 different classifiers: NB with bagging and boosting, NBd (NB with discrete inputs) with bagging and boosting, TAN, SSS, Decision Tree inducers like ID3 with bagging and boosting, C4.5, MC4 with bagging and boosting, OC1, SVM, kNN with bagging and boosting, PEBLS (Parallel Exemplar Based Learning System), CN2 (Direct Rule Induction algorithm) and Perceptrons. For the detailed evaluation results, interested readers can refer to [21].

## 10. The 6 Prototypic Expressions

Let us take a look at the 6 prototypic expressions: happiness, sadness, anger, surprise, disgust and fear. When compared to the many other possible expressions, these six are the easiest to recognize. Studies on these six expressions and the observations from the surveyed papers brings out many interesting points out of which three major points are covered in the paragraphs below: 1) A note on the confusions that occur when recognizing the prototypic expressions, 2) A note on the elicitation of these expressions and 3) The effect of occlusion on the recognition of each expression.

Are all the six expressions mutually distinguishable or is there an element of confusion between them? Sebe et al. have written about the 1978 study done by Ekman and Friesen which shows that anger is usually confused with disgust and fear is usually confused with surprise (page 196 from [58]). They attribute these confusions to the fact that these expressions share common facial movements and actions. But do the results of the confusion related studies of behavioral psychologists like Ekman and Friesen apply to modern day automatic face expression recognition systems as well? It is interesting to know that a study of the results (the published confusion matrices) from many of the surveyed papers do show similar confusion between anger and disgust [14], [19], [12], [21], [22], [23], [25]. However the confusion between fear and surprise was not very evident. Rather the systems confused fear with happiness [19], [21], [22], [23], [25] and fear with anger [14], [12], [22], [25]. Also, sadness was confused with anger [19], [21], [22], [23], [25]. When it comes to the ease of recognition, it has been shown that, out of the six prototypic expressions, surprise and happiness are the easiest to recognize [37], [14].

Let us now look at the elicitation of these 6 expressions. The most common approach taken by the researchers to elicit natural expressions is by showing the subjects emotion inducing films and movie clips. In the section *"Emotion Elicitation Using Films"*, Coan and Allen give a detailed note on the different films that have been used by various researchers to elicit spontaneous expressions in the subjects (page 21 from [30]). As we have seen in section 7, Sebe et al. have also used emotion inducing films to capture data for their spontaneous expression database. While doing so, they have reported that eliciting sadness and fear in the subjects was difficult [21]. However Coan and Allen's study reports that while eliciting fear was difficult (as noted by Sebe et al.), eliciting sadness was not difficult [30]. I think this difference in the reports is because of the different videos that were used to elicit sadness. Coan and Allen also reported that the emotions that they found difficult to elicit were anger and fear with anger being the toughest. They suggest that this may be due to the fact that the display of anger requires a very personal involvement which is very difficult to elicit using movie clips [30]. For the names of the various movie clips that have been used, the interested reader can refer page 21 from [30].

Let us conclude this section by studying the effect of occlusion on each of the prototypic expressions. In their occlusion related studies, Kotsia et al. have categorically listed the effect of occlusion on each of the six prototypic expressions [25]. Their studies show that occlusion of the mouth leads to inaccuracies in the recognition of anger, fear, happiness and sadness whereas the occlusion of the eyes (and brows) leads to a dip in the recognition accuracy of disgust and surprise [25].

## 11. Challenges and Future Work

The face expression research community is shifting its focus to the recognition of spontaneous expressions. As discussed earlier, the major challenge that the researchers face is the non-availability of spontaneous expression data. Capturing spontaneous expressions on images and video is one of the biggest challenges ahead. As noted by Sebe et al., if the subjects become aware of the recording and data capture process, their expressions immediately loses its authenticity [21]. To overcome this they used a hidden camera to record the subject's expressions and later asked for their consents. But moving ahead, researchers will need data that has subjects showing spontaneous expressions under different lighting and occlusion conditions. For such cases, the hidden camera approach may not work out. Asking subjects to wear scarves and goggles (to provide mouth and eye occlusions) while making them watch emotion inducing videos and subsequently varying the lighting conditions is bound to make them suspicious about the recording that is taking place which in turn will immediately cause their expressions to become unnatural or semi-authentic. Although building a truly authentic expression database (one where the subjects are not aware of the fact that their expressions are being recorded) is extremely challenging, a semi-authentic expression database (one where the subjects watch emotion eliciting videos but are aware that they are being recorded) can be built fairly easily. One of the best efforts in recent years in this direction is the creation of the MMI database. Along with posed expressions, spontaneous expressions have also been included. Furthermore, the DB is web-based, searchable and downloadable.

The most common method that has been used to elicit emotions in the subjects is by the use of emotion inducing videos and film clips. While eliciting happiness and amusement is quite simple, eliciting fear and anger is the challenging part. As mentioned in the previous section, Coan and Allen point out that the subjects need to become personally involved in order to display anger and fear. But this rarely happens when watching films and videos. The challenge lies in finding out the best alternative ways to capture anger and fear or in creating (or searching for) video sequences that are sure to anger a person or induce fear in him.

Another major challenge is labeling of the data that is available. Usually it is easy to find a bulk of unlabeled data. But labeling the data or augmenting it with meta-information is a very time consuming process and possibly error-prone. It requires expertise on the part of the observer or the AU coder. A solution to this problem is the use of semi-supervised learning that allows us to work with both labeled and unlabeled data [11].

Apart from the six prototypic expressions there are a host of other expressions that can be recognized. But capturing and recognizing spontaneous non-basic expressions is even more challenging than capturing and recognizing spontaneous basic expressions. This is still an open topic and no work seems to have been done on the same.

Currently all expressions are not being recognized with the same accuracy. As seen in the previous sections, anger is often confused with disgust. Also, an inspection of the surveyed results shows that the recognition percentages are different for different expressions. Looking forward, researchers must work towards eliminating such confusions and recognize all the expressions with equal accuracy.

As discussed in the previous sections, differences do exist in facial features and facial expressions between cultures (for example, Europeans and Asians) and age groups (adults and children). Face expression recognition systems must become robust against such changes.

There is also the aspect of temporal dynamics and timings. Not much work has been done on exploiting the timing and temporal dynamics of the expressions in order to differentiate between posed and spontaneous expressions. As seen in the previous sections, it has been shown from psychological studies that it is useful to include the temporal dynamics while recognizing faces since expressions not only vary in their characteristics but they vary also in their onset, apex and offset timings.

Another area where more work is required is the automatic recognition of expressions and AUs from different head angles and rotations. As seen in section 8, Pantic and Rothkrantz have worked on expression recognition from profile faces. But there seems to be no published work on recognizing expressions from faces that are at different angles between full-frontal and full-profile.

Many of the systems still require manual intervention. For the purpose of tracking the face, many systems require facial points to be manually located on the first frame. The challenge is to make the system fully automatic. In recent years, there have been advances in building such fully automatic face expression recognizers [10], [37], [13].

Looking forward, researchers can also work towards the fusion of expression-analysis and expression-synthesis. This is possible given the recent advances in animation technology, the arrival of the MPEG-4 FA standards and the available mappings of the FACS AUs to MPEG-4 FAPs.

A possible work for the future is the automatic recognition of microexpressions. Currently training kits are available that can train a human to recognize microexpressions. It will be interesting to see the kind of training that the machine will need.

I will conclude this section with a small note on the medical conditions that lead to *'loss of facial expression'*. Although it has no direct implication to the current state of the art, I feel that, as a face expression recognition researcher, it is important to know that there exist certain conditions which can cause what is known as *'flat affect'* or the condition where a person is unable to display facial expressions [87]. There are 12 causes for the loss of facial expressions namely: Asperger syndrome, Autistic disorder, Bell's palsy, Depression, Depressive disorders, Facial paralysis, Facial weakness, Hepatolenticular degeneration, Major depressive disorder, Parkinson's disease, Scleroderma, and Wilson's disease [87].

## 12. Conclusion

This paper's objective was to introduce the recent advances in face expression recognition and the associated areas in a manner that should be understandable even by the new comers who are interested in this field but have no background knowledge on the same. In order to do so, we have looked at the various aspects of face expression recognition in detail. Let us now summarize: We started with a time-line view of the various works on expression recognition. We saw some applications that have been implemented and other possible areas where automatic expression recognition can be applied. We then looked at facial parameterization using FACS AUs and MPEG-4 FAPs. Then we looked at some notes on emotions, expressions and features followed by the characteristics of an ideal system. We then saw the recent advances in face detectors and trackers. This was followed by a note on the databases followed by a summary of the state of the art. A note on classifiers was presented followed by a look at the six prototypic expressions. The last section was the challenges and possible future work to be done.

Face expression recognition systems have improved a lot over the past decade. The focus has definitely shifted from posed expression recognition to spontaneous expression recognition. The next decade will be interesting since I think that robust spontaneous expression recognizers will be developed and deployed in real-time systems and used in building emotion sensitive HCI interfaces. This is going to have an impact on our day to day life by enhancing the way we interact with computers or in general, our surrounding living and work spaces.

---